\documentclass[runningheads]{llncs}

\usepackage{multirow}
\usepackage{array,booktabs}
\usepackage{mathtools}
\usepackage{capt-of}

\usepackage{graphicx}
\usepackage{amsmath,amssymb} 
\usepackage{color}
\usepackage[width=122mm,left=12mm,paperwidth=146mm,height=193mm,top=12mm,paperheight=217mm]{geometry}

\newcommand{\smallsec}[1]{{\bf #1.}}

\DeclareMathOperator*{\argmax}{arg\,max}

\newcommand{\mI}{\mathcal{I}}
\newcommand{\mF}{\mathcal{O}}

\newcommand{\mL}{\mathcal{L}}
\newcommand{\Lp}{L}

\newcommand{\Lm}{L'}
\newcommand{\mIs}{\mI_{s}}

\newcommand{\Loss}{\mL}
\newcommand{\Lossfs}{\Loss_{pix}}
\newcommand{\Lossp}{\Loss_{point}}

\newcommand{\Lossimg}{\Loss_{img}}
\newcommand{\Lossobj}{\Loss_{obj}}

\newcommand{\pointclss}{\pointcls{ }}
\newcommand{\squiggleclss}{\squigglecls{ }}
\newcommand{\pointinsts}{\pointinst{ }}
\newcommand{\pointcls}{$1Point$}
\newcommand{\squigglecls}{$1Squiggle$}
\newcommand{\pointinst}{$AllPoints$}
\newcommand{\IMG}{$Img${ }}
\newcommand{\OBJ}{$Obj${ }}

\newcommand{\OBJns}{$Obj$}

\begin{document}
\pagestyle{headings}
\mainmatter

\title{What's the Point: \\Semantic Segmentation with Point Supervision} 

\titlerunning{What's the Point: Semantic Segmentation with Point Supervision}

\authorrunning{Amy Bearman, Olga Russakovsky, Vitto Ferrari, Li Fei-Fei}

\author{
Amy Bearman$^1$
\and
Olga Russakovsky$^2$
\and
Vittorio Ferrari$^3$
\and
Li Fei-Fei$^1$
}

\institute{Stanford University \\
\email{ \{abearman,feifeili\}@cs.stanford.edu} 
\and 
	Carnegie Mellon University \\
	\email{ olgarus@cmu.edu}
\and
	University of Edinburgh \\
	\email{ vittorio.ferrari@ed.ac.uk} 
}

\maketitle

\begin{abstract}
The semantic image segmentation task presents a trade-off between test time accuracy and training-time annotation cost. Detailed per-pixel annotations enable training accurate models but are very time-consuming to obtain; image-level class labels are an order of magnitude cheaper but result in less accurate models. We take a natural step from image-level annotation towards stronger supervision: we ask annotators to \emph{point} to an object if one exists. We incorporate this point supervision along with a novel objectness potential in the training loss function of a CNN model. Experimental results on the PASCAL VOC 2012 benchmark reveal that the combined effect of point-level supervision and objectness potential yields an improvement of $12.9\%$ mIOU over image-level supervision. Further, we demonstrate that models trained with point-level supervision are more accurate than models trained with image-level, squiggle-level or full supervision given a fixed annotation budget. 

\end{abstract}

\section{Introduction}

At the forefront of visual recognition is the question of how to effectively teach computers new concepts. Algorithms trained from carefully annotated data enjoy better performance than their weakly supervised counterparts (e.g.,~\cite{Girshick14} vs.~\cite{TouchCut},~\cite{Clark05} vs.~\cite{papandreouweakly},~\cite{Long15} vs.~\cite{pathak2015constrained}), yet obtaining such data is very time-consuming~\cite{Long15,ILSVRC}.

\begin{figure}[t]
\begin{minipage}{0.78\linewidth}
\centering
\newcommand{\w}{0.32\linewidth}
\setlength{\tabcolsep}{4pt}
\begin{tabular}{m{2.6cm} m{2.6cm} m{2.6cm}}
\centering {\bf  {\scriptsize Original image}} & 
\multicolumn{1}{c}{\scriptsize Image-level labels} &
\multicolumn{1}{c}{\scriptsize 1 point per class} \\
\centering \includegraphics[width=0.75in]{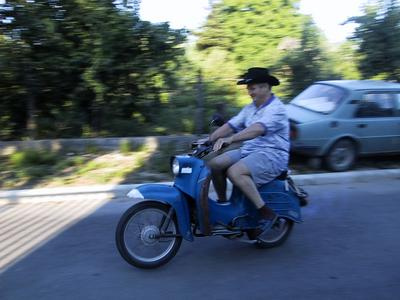} &
\centering \includegraphics[width=0.75in]{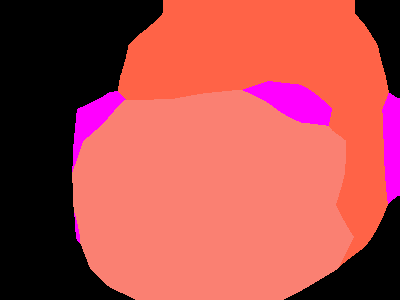} &
\centering \includegraphics[width=0.75in]{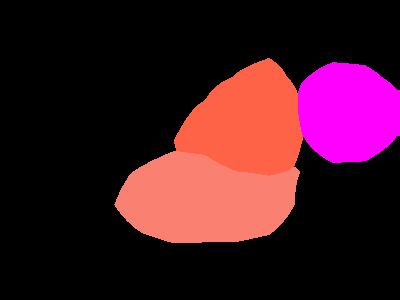} 
\end{tabular}
\end{minipage}
\begin{minipage}{0.18\linewidth}
\centering
\begin{tabular}{|l}
{\tiny Legend}\\
\includegraphics[height=0.1in]{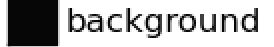}\\
\includegraphics[height=0.1in]{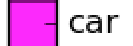}\\
\includegraphics[height=0.1in]{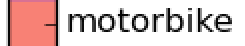}\\
\includegraphics[height=0.1in]{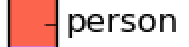}
\end{tabular}
\end{minipage}
\caption{Semantic segmentation models trained with our point-level supervision are much more accurate than models trained with image-level supervision (and even more accurate than models trained with full pixel-level supervision given the same annotation budget). The second two columns show test time results.}
\label{fig:pull-fig}
\end{figure}

It is particularly difficult to collect training data for semantic segmentation, i.e., the task of assigning a class label to every pixel in the image. Strongly supervised methods require a training set of images with per-pixel annotations~\cite{Clark05,Simonyan14,Merrill07,Hild03,Farabet13,Gould:CVPR2012} (Fig.~\ref{fig:pull-fig}). Providing an accurate outline of a single object takes between $54$ seconds~\cite{Jain13} and $79$ seconds~\cite{Long15}. A typical indoor scene contains $23$ objects~\cite{Guillaumin15}, raising the annotation time to tens of minutes per image. Methods have been developed to reduce
the annotation time through effective interfaces~\cite{Long15,GrabCut,COCO,Xu15,ScribbleSup,bell15minc}, e.g., through requesting human feedback only as necessary~\cite{Jain13}. Nevertheless, accurate per-pixel annotations remain costly and scarce.

To alleviate the need for large-scale detailed annotations, weakly supervised semantic segmentation techniques have been developed. The most common setting is where only image-level labels for the presence or absence of classes are provided during training~\cite{papandreouweakly,Vezhnevets11,Vezhnevets12,Song14,Pathak15,Xu14,Pinheiro15}, but other forms of weak supervision have been explored as well, such as bounding box annotations~\cite{papandreouweakly}, eye tracks~\cite{Papadopoulos14}, free-form squiggles~\cite{Xu15,ScribbleSup}, or noisy web tags~\cite{Ahmed14}. These methods require significantly less annotation effort during training, but are not able to segment new images nearly as accurately as fully supervised techniques. 

In this work, we take a natural step towards stronger supervision for semantic segmentation at negligible additional time, compared to image-level labels. The most natural way for humans to refer to an object is by pointing: ``That cat over there" \emph{(point)} or ``What is that over there?" \emph{(point)}. Psychology research has indicated that humans point to objects in a consistent and predictable way~\cite{Clark05,Firestone14}. The fields of robotics~\cite{Hild03,Sauppe14} and human-computer interaction~\cite{Merrill07} have long used pointing as the effective means of communication. However, point annotation is largely unexplored in semantic segmentation.

Our \textbf{primary contribution} is a novel supervision regime for semantic segmentation based on humans pointing to objects. We extend a state-of-the-art convolutional neural network (CNN) framework for semantic segmentation~\cite{Long15,Pathak15} to incorporate point supervision in its training loss function. With just one annotated point per object class, we considerably improve semantic segmentation accuracy. We ran an extensive human study to collect these points on the PASCAL VOC 2012 dataset and evaluate the annotation times. We also make the user interface and the annotations available to the community. 

One lingering concern with supervision at the point level is that it is difficult to infer the full extent of the object. Our \textbf{secondary contribution} is incorporating an generic objectness prior~\cite{Alexe12} directly in the loss to guide the training of a CNN. This prior helps separate objects (e.g., car, sheep, bird) from background (e.g., grass, sky, water), by providing a probability that a pixel belongs to an object. Such priors have been used in segmentation literature for selecting image regions to segment~\cite{Bharath14}, as unary potentials in a conditional random field model~\cite{Vezhnevets11}, or during inference~\cite{Pinheiro15}. However, to the best of our knowledge, we are the first to employ this directly in the loss to guide the training of a CNN. 

The combined effect of our contributions is a substantial increase of $12.9\%$ mean intersection over union (mIOU) on the PASCAL VOC 2012 dataset~\cite{PASCALIJCV} compared to training with image-level labels. Further, we demonstrate that models trained with point-level supervision outperform models trained with image-level, squiggle-level, and full supervision by $2.7-20.8\%$ mIOU given a fixed annotation budget.

\section{Related Work}
\label{sec:related_work}

\smallsec{Types of Supervision for Semantic Segmentation} 
To reduce the up-front annotation time for semantic segmentation, recent works have focused on training models in a weakly- or semi-supervised setting. Many forms of supervision have been explored, such as eye tracks~\cite{Papadopoulos14}, free-form squiggles~\cite{Xu15,ScribbleSup}, noisy web tags~\cite{Ahmed14}, size constraints on objects~\cite{pathak2015constrained} or heterogeneous annotations~\cite{hong2015decoupled}. Common settings are image-level labels~\cite{papandreouweakly,Pathak15,Pinheiro15} and bounding boxes~\cite{papandreouweakly,dai2015boxsup}.~\cite{Guillaumin15,Chai11,Joulin10} use co-segmentation methods trained from image-level labels to automatically infer the segmentations.~\cite{pathak2015constrained,Pathak15,Pinheiro15} train CNNs supervised only with image-level labels by extending the Multiple-Instance Learning (MIL) framework for semantic segmentation.~\cite{papandreouweakly,dai2015boxsup} use an EM procedure, which alternates between estimating pixel labels from bounding box annotations and optimizing the parameters of a CNN.

There is a trade-off between annotation time and accuracy: models trained with higher levels of supervision perform far better than weakly-supervised models, but require large strongly-supervised datasets, which are costly and scarce. We propose an intermediate form of supervision, using points, which adds negligible additional annotation time to image-level labels, yet achieves better results.~\cite{bell15minc} also uses point supervision during training, but it trains a patch-level CNN classifier to serve as a unary potential in a CRF, whereas we use point supervision directly during CNN training.

\smallsec{CNNs for Segmentation}
Recent successes in semantic segmentation have been driven by methods that train CNNs originally built for image classification to assign semantic labels to each pixel in an image~\cite{Long15,Farabet13,Bharath14,DeepLab-strong}. One extension of the fully convolutional network (FCN) architecture developed by~\cite{Long15} is to train a multi-layer deconvolution network end-to-end~\cite{noh2015learning}. More inventive forms of post-processing have also been developed, such as combining the responses at the final layer of the network with a fully-connected CRF~\cite{DeepLab-strong}. We develop our approach on top of the basic framework common to many of these methods. 

\smallsec{Interactive Segmentation}
Some semantic segmentation methods are interactive, in that they collect additional annotations at test time to refine the segmentation. These annotations can be collected as points~\cite{TouchCut} or free-form squiggles~\cite{GrabCut}. These methods require additional user input at test time; in contrast, we only collect user points once and only use them at training time.

\section{Semantic Segmentation Method}
\label{sec:method}

\begin{figure}[t]
\centering
\begin{tabular}{ccc}
Original image & FCN~\cite{Long15} & Segmentation \\
\includegraphics[width=0.2\linewidth]{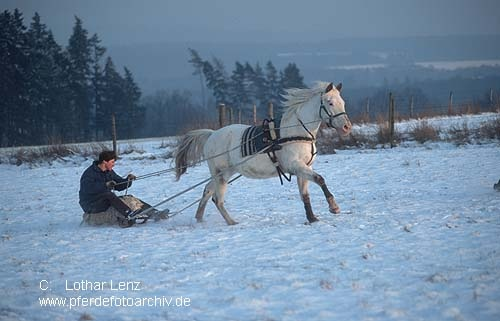} &
\includegraphics[width=0.2\linewidth]{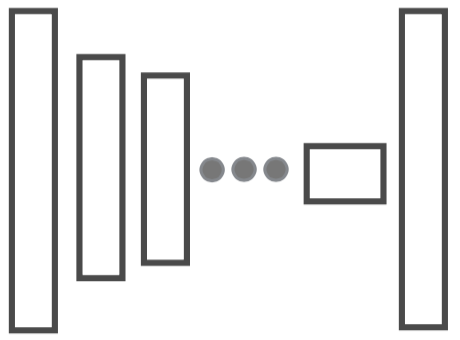} & 
\includegraphics[width=0.2\linewidth]{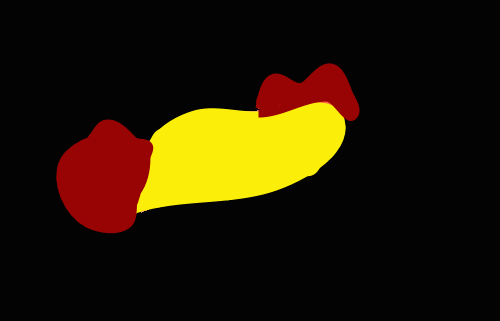} \\
\end{tabular}
\begin{tabular}{cccc}
\multicolumn{4}{c}{{\bf Levels of supervision}} \\
\includegraphics[width=0.2\linewidth]{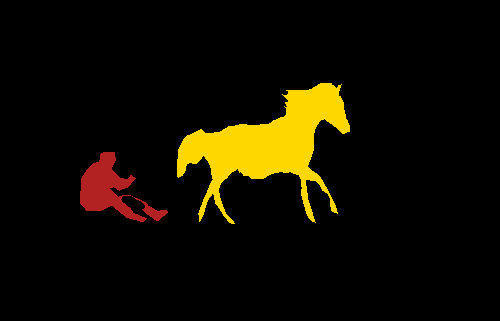}  &
\includegraphics[width=0.2\linewidth]{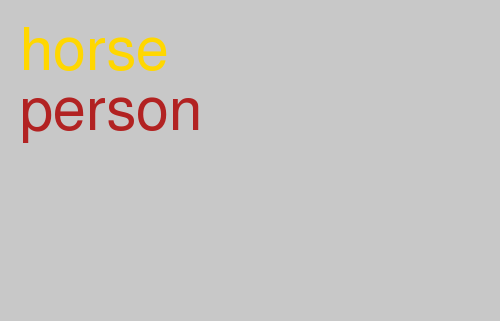}   &
\includegraphics[width=0.2\linewidth]{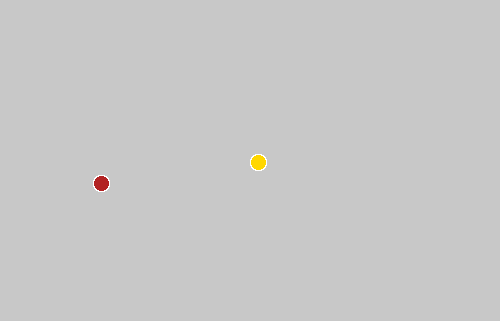}  &
\includegraphics[width=0.2\linewidth]{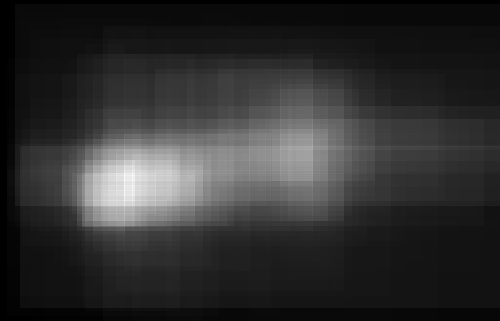}   \\
Full& 
Image-level&
Point-level &
Objectness prior \\
\end{tabular}
\caption{(\emph{Top}): Overview of our semantic segmentation training framework. (\emph{Bottom}): Different levels of training supervision. For full supervision, the class of every pixel is provided. For image-level supervision, the class labels are known but their locations are not. We introduce point-level supervision, where each class is only associated with one or a few pixels, corresponding to humans pointing to objects of that class. We include an objectness prior in our training loss function to accurately infer the object extent.}
\label{fig:framework}
\end{figure}

We describe here our approach to using point-level supervision (Fig.~\ref{fig:framework}) for training semantic segmentation models. In Section~\ref{sec:human}, we will demonstrate that this level of supervision is cheap and efficient to obtain. In our setting (in contrast to~\cite{TouchCut}), supervised points are only provided on training images. The learned model is then used to segment test images with no additional human input.

Current state-of-the-art semantic segmentation methods~\cite{papandreouweakly,Long15,Pathak15,Pinheiro15,DeepLab-strong}, both supervised and unsupervised, employ a unified CNN framework. These networks take as input an image of size $W \times H$ and output a $W \times H \times N$ score map where $N$ is the set of classes the CNN was trained to recognize (Fig.~\ref{fig:framework}). At test time, the score map is converted to per-pixel predictions of size $W \times H$ by either simply taking the maximally scoring class at each pixel~\cite{Long15,Pathak15} or employing more complicated post-processing~\cite{papandreouweakly,Pinheiro15,DeepLab-strong}. 

Training models with different levels of supervision requires defining appropriate loss functions in each scenario. We begin by presenting two of the most commonly used in the literature. We then extend them to incorporate (1) our proposed point supervision and (2) a novel objectness prior.

\smallsec{Full Supervision}
When the class label is available for every pixel during training, the CNN is commonly trained by optimizing the sum of per-pixel cross-entropy terms~\cite{Long15,DeepLab-strong}. Let $\mI$ be the set of pixels in the image. Let $s_{ic}$ be the CNN score for pixel $i$ and class $c$. Let $S_{ic} = \exp(s_{ic})/\sum_{k=1}^N \exp(s_{ik})$ be the softmax probability of class $c$ at pixel $i$. Given a ground truth map $G$ indicating that pixel $i$ belongs to class $G_i$, the loss on a single training image is:
\begin{equation}
\label{eq:lossfs}
\Lossfs(S,G) = - \sum_{i \in \mI} \log(S_{iG_i}) 
\end{equation}
The loss is simply zero for pixels where the ground truth label is not defined (e.g., in the case of pixels defined as ``difficult'' on the boundary of objects in PASCAL VOC~\cite{PASCALIJCV}).

\smallsec{Image-Level Supervision}
In this case, the only information available during training are the sets $\Lp \subseteq \{1, \dots , N\}$ of classes present in the image and $\Lm \subseteq \{1, \dots , N\}$ of classes not present in the image. The CNN model can be trained with a different cross-entropy loss:
\begin{align}
\label{eq:lossimg}
\Lossimg(S,\Lp,\Lm) &=  - \frac{1}{|\Lp|} \sum_{c \in L} \log(S_{{t_c}c}) - \frac{1}{|\Lm|} \sum_{c \in \Lm}\log(1-S_{t_cc}) \\
&\mbox{with } t_c = \argmax_{i \in \mI} S_{ic}
\notag
\end{align}
The first part of Eqn.~\eqref{eq:lossimg}, corresponding to $c \in L$, is used in~\cite{Pathak15}.
It encourages each class in $L$ to have a high probability on at least one pixel in the image. The second part has been added in~\cite{pathak2015constrained}, corresponding to the fact that no pixels should have high probability for classes that are not present in the image.

\smallsec{Point-Level Supervision}
We study the intermediate case where the object classes are known for a small set of supervised pixels $\mIs$, whereas other pixels are just known to belong to some class in $\Lp$.  We generalize Eqns.~\eqref{eq:lossfs} and \eqref{eq:lossimg} to:
\small
\begin{equation}
\label{eq:lossp}
\Lossp(S,G,\Lp,\Lm) = \Lossimg(S,\Lp,\Lm) -  \sum_{i \in \mIs} \alpha_i \log(S_{iG_i})   
\end{equation}
\normalsize
Here, $\alpha_i$ determines the relative importance of each supervised pixel. We experiment with several formulations for $\alpha_i$.
(1), for each class we ask the user to either determine that the class is not present in the image or to point to one object instance. In this case, $|\mIs| = |L|$ and $\alpha_i$ is uniform for every point;
(2), we ask multiple annotators to do the same task as (1), and we set $\alpha_i$ to be the confidence of the accuracy of the annotator that provided the point;
(3), we ask the annotator(s) to point to every \emph{instance} of the classes in the image, and $\alpha_i$ corresponds to the \emph{order} of the points: the first point is more likely to correspond to the largest object instance and thus deserves a higher weight $\alpha_i$.

\smallsec{Objectness Prior}
One issue with training models with very few or no supervised pixels is correctly inferring the spatial extent of the objects. In general, weakly supervised methods are prone to local minima: focusing on only a small part of the target object, or predicting all pixels as belonging to the background class~\cite{Pathak15}. To alleviate this problem, we introduce an additional term in our training objective based on an objectness prior (Fig.~\ref{fig:framework}). Objectness provides a probability for whether each pixel belongs to {\em any} object class \cite{Alexe12} (e.g., bird, car, sheep), as opposed to background (e.g., sky, water, grass).
These probabilities have been used in the weakly supervised
semantic segmentation literature before as unary potentials in graphical models~\cite{Vezhnevets11} or during inference following a CNN segmentation~\cite{Pinheiro15}. To the best of our knowledge, we are the first to incorporate them directly into CNN training.

Let $P_i$ be the probability that pixel $i$ belongs to an object. Let $\mF$ be the classes corresponding to objects, with the other classes corresponding to backgrounds. In PASCAL VOC, $\mF$ is the 20 object classes, and there is a single generic background class.
We define a new loss:
\begin{equation} 
\label{eq:lossobj}
\Lossobj(S,P) = - \frac{1}{|\mI|} \sum_{i \in \mI} P_i \log \left (\sum_{c \in \mF} S_{ic} \right ) + 
(1-P_i) \log \left (1 - \sum_{c \in \mF} S_{ic} \right )
\end{equation}
At pixels with high $P_i$ values, this objective encourages placing probability mass on object classes. Alternatively, when $P_i$ is low, it prefers mass on the background class.
Note that $\Lossobj$ requires no human supervision (beyond pre-training the generic objectness detector), and thus can be combined with any loss above.

\section{Crowdsourcing Annotation Data}
\label{sec:human}

In this section, we describe our method for collecting annotations for the different levels of supervision. The annotation time required for point-level and squiggle-level supervision was measured directly during data collection. For other types of supervision, we rely on the annotation times reported in the literature.

\smallsec{Image-Level Supervision (20.0 sec/img)} Collecting image-level labels takes $1$ second per class~\cite{Papadopoulos14}. Thus, annotating an image with $20$ object classes in PASCAL VOC is expected to take $20$ seconds per image.

\smallsec{Full Supervision (239.7 sec/img)} There are $1.5$ object classes per image on average in PASCAL VOC 2012~\cite{PASCALIJCV}. It takes $1$ second to annotate every object that is not present (to obtain an image-level ``no'' label), for $18.5$ seconds of labeling time. Additionally, there are $2.8$ object instances on average per image that need to be segmented~\cite{PASCALIJCV}.  
  The authors of the COCO dataset report $22$ worker hours for 1,000 segmentations~\cite{COCO}. This implies a mean labeling time of $79$ seconds per object segmentation, adding $2.8 \times 79$ seconds of labeling in our case. Thus, the total expected annotation time is $239.7$ seconds per image.

\subsection{Point-Level Supervision (22.1 sec/img)} 
\begin{figure}[t]
\begin{minipage}{0.3\linewidth}
\centering
\includegraphics[width=0.9\linewidth]{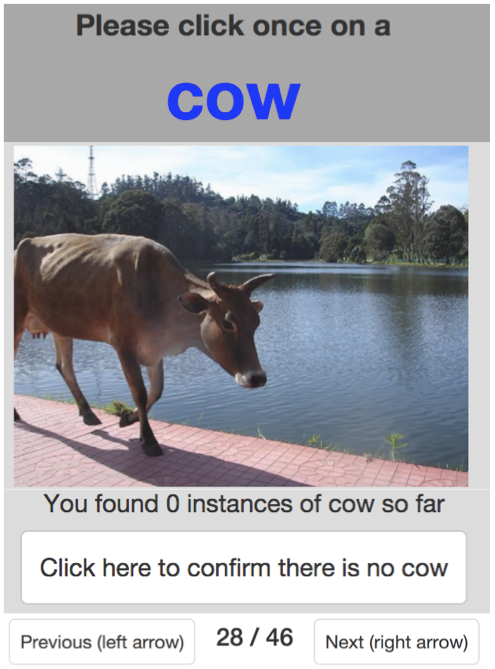}
\end{minipage}\hfill
\begin{minipage}{0.3\linewidth}
\centering
\includegraphics[width=0.9\linewidth]{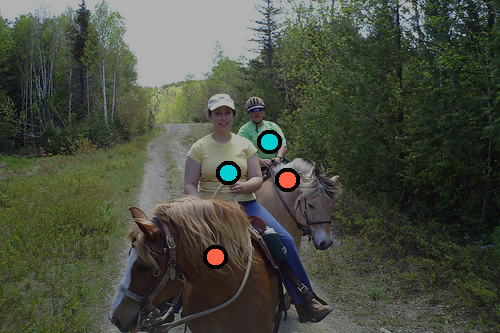}
\includegraphics[width=0.9\linewidth]{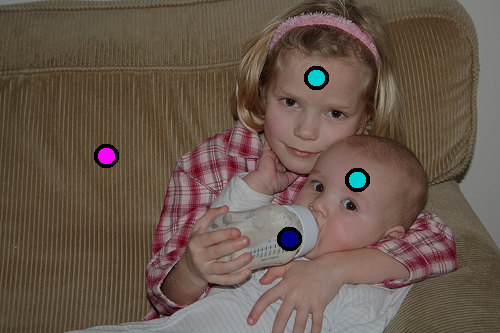}
\end{minipage}\hfill
\begin{minipage}{0.3\linewidth}
\centering
\includegraphics[width=0.9\linewidth]{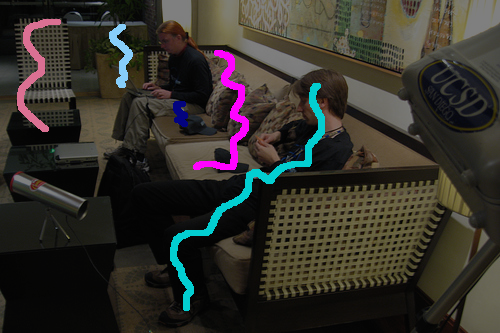}
\includegraphics[width=0.9\linewidth]{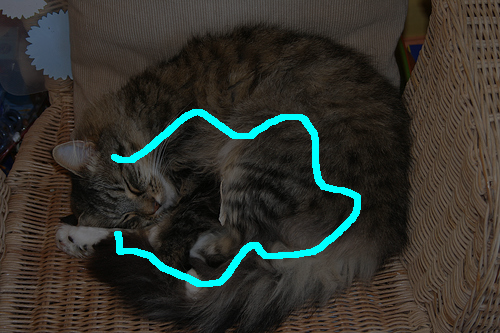}
\end{minipage}
\caption{\emph{Left.} AMT annotation UI for point-level supervision. \emph{Center.} Example points collected. \emph{Right.} Example squiggles collected. Colors correspond to different classes.}
\label{fig:amt-ui}
\end{figure}

We used Amazon Mechanical Turk (AMT) to annotate point-level supervision on 20 PASCAL VOC object classes over 12,031 images: all training and validation images of the PASCAL VOC 2012 segmentation task~\cite{PASCALIJCV} plus the additional images of~\cite{Hariharan11}. Fig.~\ref{fig:amt-ui} (left) shows the annotation inferface and Fig.~\ref{fig:amt-ui} (center) shows some collected data. We use two different point-level supervision tasks. For each image, we obtain either (1) one annotated point per object class, on the first instance of the class the annotator sees (\pointcls), and (2) one annotated point per object instance (\pointinst). We make these collected annotations and the annotation system publicly available.

\smallsec{Annotation Time} There are $1.5$ classes on average per image in PASCAL VOC 2012. It takes workers a median of $2.4$ seconds to click on the first instance of an object.  Therefore, the labeling time of \pointclss is $1 \times 18.5 + 1.5 \times 2.4 = \mathbf{22.1}$ seconds per image. 
It takes workers a median of $0.9$ seconds to click on every additional instance of an object class. There are $2.8$ instances on average per image, thus the labeling time of \pointinsts is $1 \times 18.5 + 1.5\times2.4 + (2.8-1.5) \times 0.9 = \mathbf{23.3}$ seconds per image. Note that point supervision is only $1.1$-$1.2$x more time-consuming than obtaining image-level labels, and more than $10$x cheaper than full supervision.

\smallsec{Quality Control} Quality control for point annotation was done by planting 10 evaluation images in a 50-image task and ensuring that at least 8 are labeled correctly. We consider a point correct if it falls inside a tight bounding box around the object. For the \pointinsts task, the number of annotated clicks must be at least the number of known object instances.

\smallsec{Error Rates} Simply determining the presence or absence of an object class in an image was fairly easy, and workers incorrectly labeled an object class as absent only $1.0\%$ of the time. On the \pointclss task, $7.2\%$ of points were on a pixel with a different class label (according to the PASCAL ground truth), and an additional $0.8\%$ were on an unclassified ``difficult'' pixel. For comparison,~\cite{Russakovsky15} reports much higher $25\%$ average error rates when drawing bounding boxes. Our collected data is high-quality, confirming that pointing to objects comes naturally to humans~\cite{Clark05,Merrill07}.

Annotators had more difficulty with the \pointinsts class: $7.9\%$ of ground truth instances were left unannotated, $14.8\%$ of the clicks were on the wrong object class, and $1.6\%$ on ``difficult'' pixels. This task caused some confusion among workers due to blurry or very small instances; for example, many of these instances are not annotated in the ground truth but were clicked by workers, accounting for the high false positive rate. 

\subsection{Squiggle-Level Supervision (34.9 sec/img)}

\cite{Xu15,ScribbleSup} have experimented with training with free-form squiggles, where a subset of pixels are labeled. While~\cite{Xu15} simulates squiggles by randomly labeling superpixels from the ground truth, we follow~\cite{ScribbleSup} in collecting squiggle annotations (and annotation times) from humans for 20 object classes on all PASCAL VOC 2012 trainval images. This allows us to properly compare this supervision setting to human points. We extend the user interface shown in Fig.~\ref{fig:amt-ui} (left) by asking annotators to draw one squiggle on one instance of the target class. Fig.~\ref{fig:amt-ui} (right) shows some collected data.

\smallsec{Annotation Time} As before, it takes $18.5$ seconds to annotate the classes not present in the image. For every class that is present, it takes $10.9$ seconds to draw a free-form squiggle on the target class. Therefore, the labeling time of \squiggleclss is $18.5 + 1.5 \times 10.9 = \mathbf{34.9}$ seconds per image. This is $1.6$x more time-consuming than obtaining \pointclss point-level supervision and $1.7$x more than image-level labels. 

\smallsec{Error Rates} We used similar quality control to point-level supervision. Only $6.3\%$ of the annotated pixels were on the wrong object class, and an additional $1.4\%$ were on pixels marked as ``difficult'' in PASCAL VOC~\cite{PASCALIJCV}. 

In Section~\ref{sec:experiments} we compare the accuracy of the models trained with different levels of supervision.

\section{Experiments}
\label{sec:experiments}

We empirically demonstrate the efficiency of our point-level and objectness prior. We compare these forms of supervision against image-level labels, squiggle-level, and fully supervised data. We conclude that point-level supervision makes a much more efficient use of annotator time, and produces much more effective models under a fixed time budget.

\subsection{Setup}
\smallsec{Dataset}
We train and evaluate on the PASCAL VOC 2012 segmentation dataset~\cite{PASCALIJCV} augmented with extra annotations from~\cite{Hariharan11}. There are 10,582 training images, 1,449 validation images and 1,456 test images. We report the mean intersection over union (mIOU), averaged over 21 classes.

\smallsec{CNN Architecture}
We use the state-of-the-art fully convolutional network model~\cite{Long15}. Briefly, the architecture is based on the VGG 16-layer net~\cite{Simonyan14}, with all fully connected layers converted to convolutional layers. The last classifier layer is discarded and replaced with a 1x1 convolution layer with channel dimension $N=21$ equal to the number of object classes. The final modification is the addition of a  deconvolution layer to bilinearly upsample the output to pixel-level dense predictions.

\smallsec{CNN Training} 
We train following a procedure similar to~\cite{Long15}. We use stochastic gradient descent with a fixed learning rate of $10^{-5}$, doubling the learning rate for biases, and with a minibatch of $20$ images, momentum of  $0.9$ and weight decay $0.0005$.
The network is initialized with weights pre-trained for a 1000-way classification task of the ILSVRC 2012 dataset~\cite{Long15,ILSVRC,Simonyan14}.\footnote{Standard in the literature~\cite{Girshick14,papandreouweakly,Long15,Pathak15,Pinheiro15,DeepLab-strong}. We do not consider the cost of collecting those annotations; including them would not change our overall conclusions.}
In the fully supervised case we zero-initialize the classifier weights~\cite{Long15}, and for all the weakly supervised cases we follow~\cite{Pathak15} to initialize them with weights learned by the original VGG network for classes common to both PASCAL and ILSVRC. We backpropagate through all layers to fine-tune the network, and train for 50,000 iterations. We build directly on the publicly available implementation of~\cite{Long15,caffe}.\footnote{\cite{Long15} introduces additional refinement by decreasing the stride of the output layers from 32 pixels to 8 pixels, which improves their results from $59.7\%$ to $62.7\%$ mIOU on the PASCAL VOC 2011 validation set. We use the original model with stride of 32 for simplicity.} 

\smallsec{Objectness prior} We calculate the per-pixel objectness prior by assigning each pixel the average objectness score of all windows containing it. These scores are obtained by using the pre-trained model from the released code of~\cite{Alexe12}. The model is trained on 50 images with 291 object instances randomly sampled from a variety of different datasets (e.g., INRIA Person, Caltech 101) that do not overlap with PASCAL VOC 2007-2012~\cite{Alexe12}. For fairness of comparison, we include the annotation cost of training the objectness prior. We estimate the 291 bounding boxes took $10.2$ seconds each on average to obtain~\cite{Russakovsky15}, for $49.5$ minutes of annotation. Amortized across the 10,582 PASCAL training images, using the objectness prior thus costs $\mathbf{0.28}$ {\bf seconds} of extra annotation per image.

\subsection{Synergy Between Point-Level Supervision and Objectness Prior}
\label{sec:point-sup}

\begin{figure*}[t]
\centering
\newcommand{\w}{0.185\linewidth}
\begin{tabular}{ccccc} 
Original &  Image-level &  Image-level & Point-level & Full \\
image & supervision & + objectness & + objectness & supervision \\

\includegraphics[width=\w]{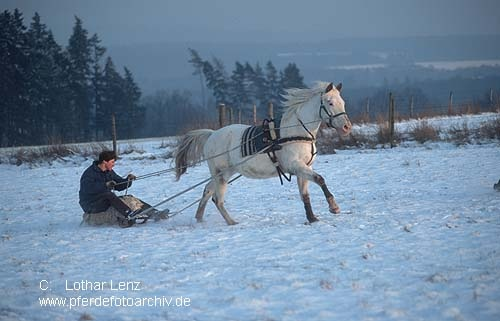} &
\includegraphics[width=\w]{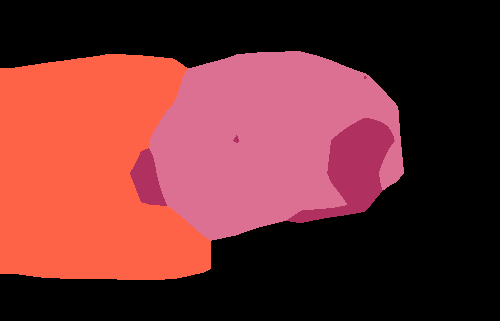} &
\includegraphics[width=\w]{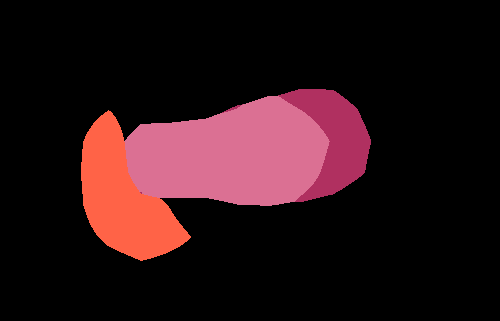} &
\includegraphics[width=\w]{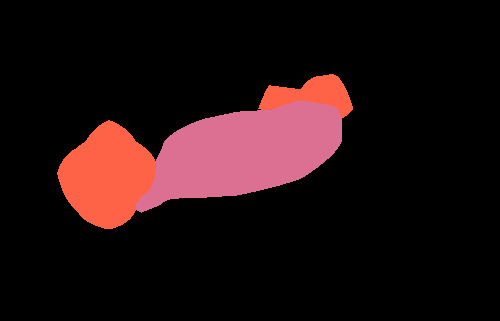} &
\includegraphics[width=\w]{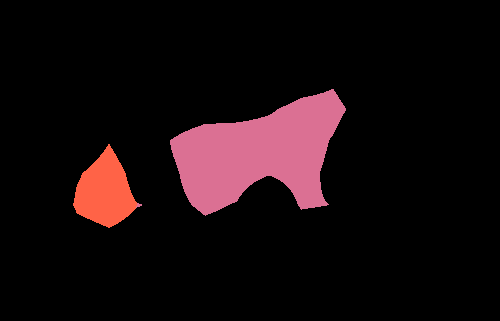}  \\
\includegraphics[width=\w]{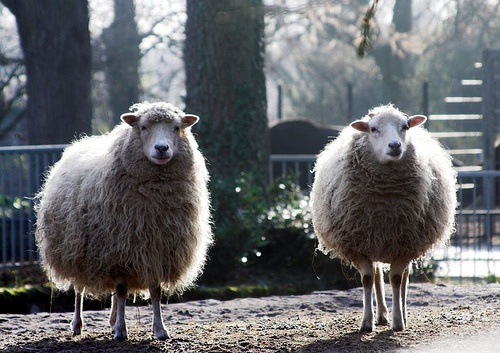} &
\includegraphics[width=\w]{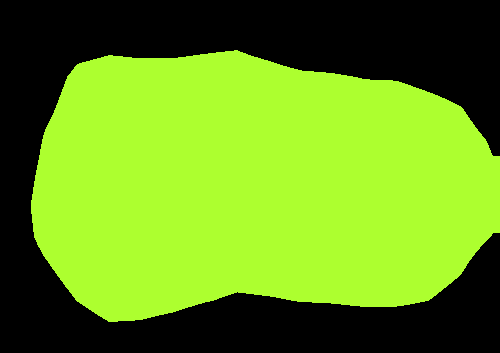} &
\includegraphics[width=\w]{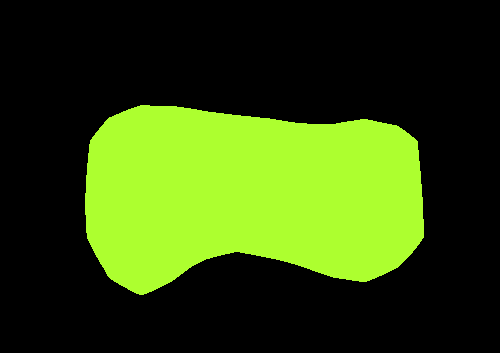} &
\includegraphics[width=\w]{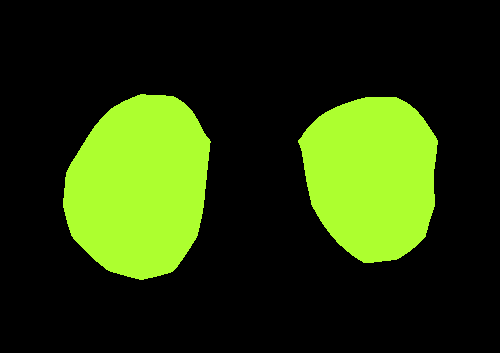} &
\includegraphics[width=\w]{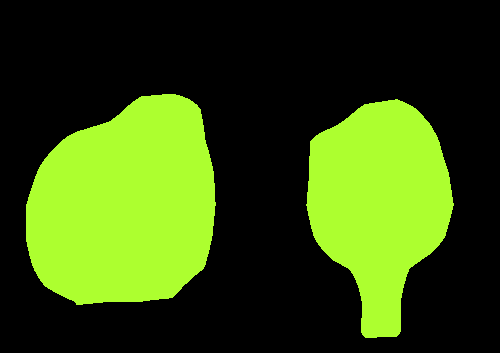} \\
\includegraphics[width=\w]{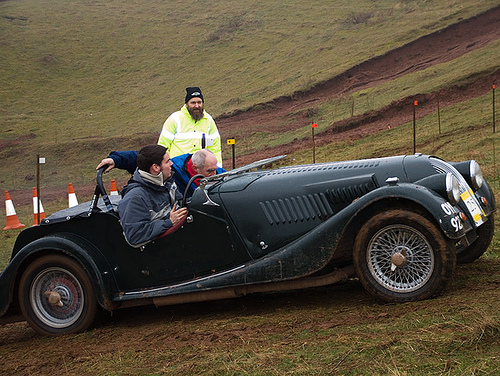} &
\includegraphics[width=\w]{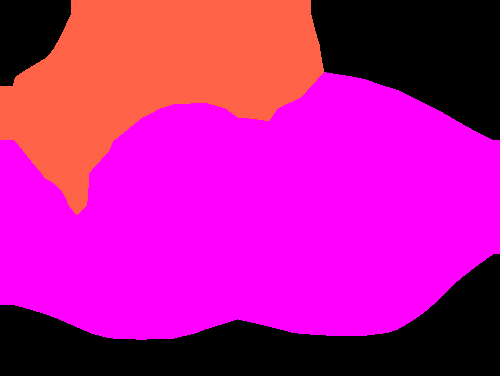} &
\includegraphics[width=\w]{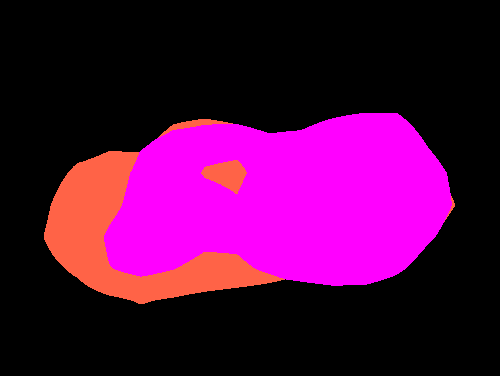} &
\includegraphics[width=\w]{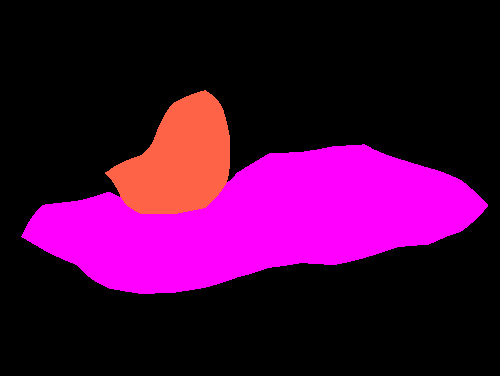} &
\includegraphics[width=\w]{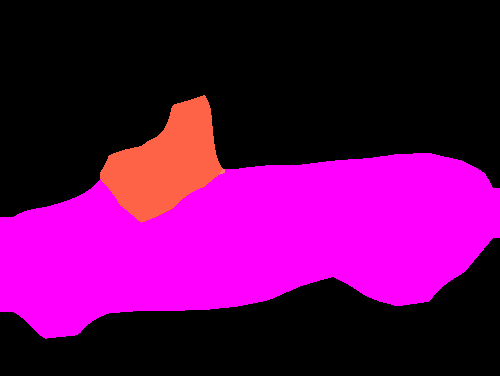} \\
\includegraphics[width=\w]{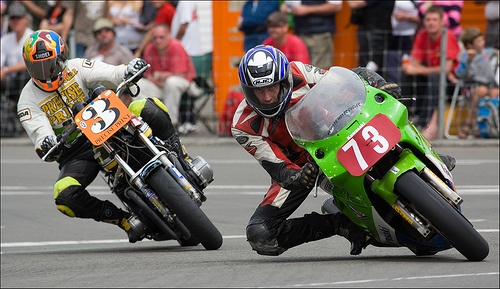} &
\includegraphics[width=\w]{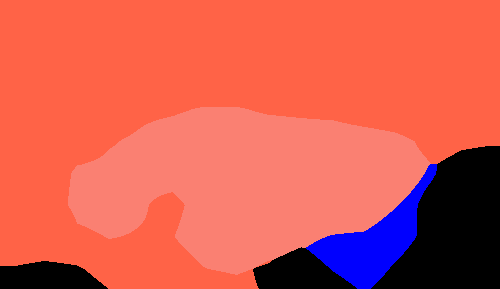} &
\includegraphics[width=\w]{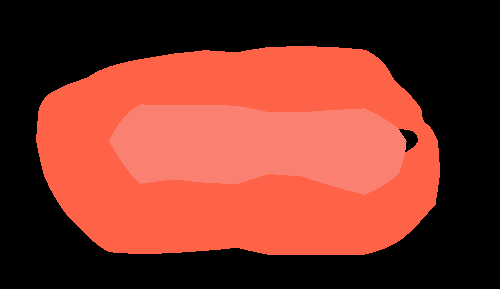} &
\includegraphics[width=\w]{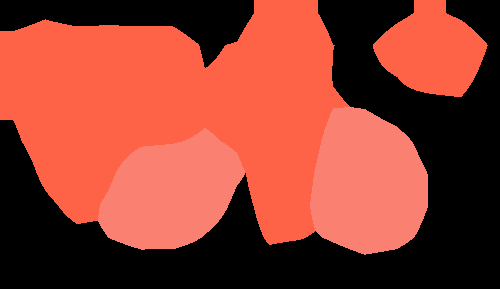} &
\includegraphics[width=\w]{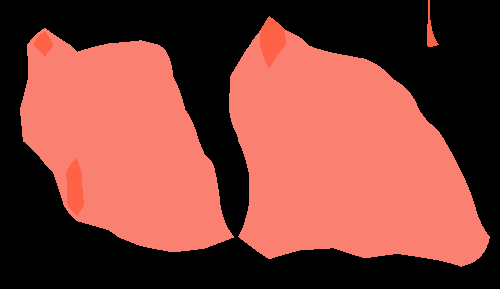} 
\end{tabular}
\newcommand{\hcb}{0.02\linewidth}
\begin{tabular}{cccccccccc}
\includegraphics[height=\hcb]{colorbar/background.png} &
\includegraphics[height=\hcb]{colorbar/car.png}&
\includegraphics[height=\hcb]{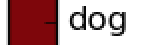}&
\includegraphics[height=\hcb]{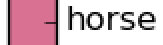}&
\includegraphics[height=\hcb]{colorbar/motorbike.png}&
\includegraphics[height=\hcb]{colorbar/person.png}&
\includegraphics[height=\hcb]{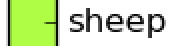}&
\end{tabular}
\caption{Qualitative results on the PASCAL VOC 2012 validation set. The model trained with image-level labels usually predicts the correct classes and their general locations, but it over-extends the segmentations. The objectness prior improves the accuracy of the image-level model by helping infer the object extent. Point supervision aids in separating distinct objects (row 2) and classes (row 4) and helps correctly localize the objects (rows 3 and 4). Best viewed in color. }
\label{fig:exs}
\end{figure*}

We first establish the baselines of our model and show the benefits of both point-level supervision and objectness prior. Table~\ref{table:pascalresults} (top) summarizes our findings and Table~\ref{table:perclass} (top) shows the per-class accuracy breakdown.

\smallsec{Baseline}
We train a baseline segmentation model from image-level labels with no additional information. We base our model on~\cite{Pathak15}, which trains a similar fully convolutional network and obtains $25.1\%$ mIOU on the PASCAL VOC 2011 validation set. We notice that the {\em absence} of a class label in an image is also an important supervisor signal, along with the presence of a class label, as in~\cite{pathak2015constrained}. We incorporate this insight into our loss function $\Lossimg$ in Eqn.~\ref{eq:lossimg}, and see a substantial $5.4\%$ improvement in mIOU from the baseline, when evaluated on the PASCAL VOC 2011 validation set.

\smallsec{Effect of Point-Level Supervision}
We now run a key experiment to investigate how having just one annotated point per class per image improves semantic segmentation accuracy. We use loss $\Lossp$ of Eqn.~\eqref{eq:lossp}. On average there are only $1.5$ supervised pixels per image (as many as classes per image). All other pixels are unsupervised. We set $\alpha = 1/n$ where $n$ is the number of supervised pixels on a particular training image.
On the PASCAL VOC 2012 validation
set, the accuracy of a model trained using $\Lossimg$ is $29.8$\%
mIOU. Adding our point supervision improves accuracy by $5.3$\% to $35.1$\% mIOU (row 3 in Table~\ref{table:pascalresults}).

\smallsec{Effect of Objectness Prior}
One issue with training models with very few or no supervised pixels is the difficulty of inferring the full extent of the object. With image-level labels, the model tends to learn that objects occupy a much greater area than they actually do (second column of Fig.~\ref{fig:exs}). 
We introduce the objectness prior in the loss using Eqn.~\eqref{eq:lossobj} to aid the model in correctly predicting the extent of objects (third column on Fig.~\ref{fig:exs}). This improves segmentation accuracy: when supervised only with image-level labels, the \IMG model obtained $29.8\%$ mIOU, and the \IMG  + \OBJ model improves to  $32.2\%$ mIOU.

\smallsec{Effect of Combining Point-Level Supervision and Objectness}
The effect of the objectness prior is even more apparent when used together with point-level supervision. When supervised with \pointcls, the \IMG model achieves $35.1\%$ mIOU, and the \IMG + \OBJ model improves to $42.7\%$ mIOU (rows 3 and 4 in Table~\ref{table:pascalresults}).
Conversely, when starting from the \IMG + \OBJ image-level model, the effect of a single point of supervision is stronger. Adding just one point per class improves accuracy by $10.5\%$ from $32.2\%$ to $42.7\%$.

\smallsec{Conclusions} We make two conclusions. First, the objectness prior is very effective for training these models with none or very few supervised pixels -- and this comes with no additional human supervision cost on the target dataset. For the rest of the experiments, whenever not all pixels are labeled (i.e., all but full supervision) we always use \IMG + \OBJ together. Second, our two contributions operate in synergetic ways. The combined effect of both point-level supervision and objectness prior is a $+13\%$ improvement (from $29.8\%$ to $42.7\%$ mIOU).

\begin{table}[t]
\centering
\setlength{\tabcolsep}{8pt}
\begin{tabular}{lclc} 
{\bf Supervision}& {\bf Time (s)} & {\bf Model} & {\bf mIOU (\%)} \\
\hline
Image-level labels & 20.0 & \IMG & 29.8 \\
Image-level labels & 20.3 & \IMG + \OBJ & 32.2 \\
\pointcls &  22.1 & \IMG & 35.1 \\
\pointcls & 22.4 & \IMG + \OBJ &  42.7 \\
\hline
\pointinst & 23.6 & \IMG + \OBJ & 42.7 \\
\pointinsts (weighted) & 23.5 & \IMG + \OBJ & 43.4 \\
\pointclss (3 annotators) & 29.6 & \IMG + \OBJ & 43.8 \\
\pointclss (random annotators) & 22.4 & \IMG + \OBJ & 42.8 - 43.8 \\
\pointclss (random points) & 240 & \IMG + \OBJ & 46.1 \\
\hline
Full supervision & 239.7 & \IMG & 58.3 \\
Hybrid approach & 24.5 & \IMG + \OBJ & 53.1 \\
1 squiggle per class & 35.2 & \IMG + \OBJ & 49.1 \\
\hline
\end{tabular}
\caption{Results on the PASCAL VOC 2012 validation set, including both annotation time (second column) and accuracy of the model (last column). Top, middle and bottom correspond to Sections~\ref{sec:point-sup}, \ref{sec:point-variations} and \ref{sec:stronger} respectively.}
\label{table:pascalresults}
\end{table}

\subsection{Point-Level Supervision Variations}
\label{sec:point-variations}

Our goal in this section is to build a deeper understanding of the properties of point-level supervision that make it an advantageous form of supervision. Table~\ref{table:pascalresults} summarizes our findings and Table~\ref{table:perclass} shows the per-class accuracy breakdown.

\smallsec{Multiple Instances}
Using points on all instances (\pointinst) instead of just one point per class  (\pointcls) remains at $42.7\%$ mIOU:  the benefit from extra supervision is offset by the confusion introduced by some difficult instances that are annotated. We introduce a weighting factor $\alpha_i = 1/2^r$ in Eqn.~\eqref{eq:lossp} where $r$ is the ranked order of the point (so the first instance of a class gets weight 1, the second instance gets weight 1/2, etc.). This \pointinsts(weighted) method improves results by a modest $0.7\%$ to $43.4\%$ mIOU. 

\smallsec{Patches} The segmentation model effectively enforces spatial label smoothness, so increasing the area of supervised pixels by a radius of 2, 5 and 25 pixels around a point has
little effect, with $43.0-43.1\%$ mIOU (not shown in Table~\ref{table:pascalresults}).

\smallsec{Multiple Annotators}
We also collected \pointclss data from 3 different annotators and used all points during training. This achieved a modest improvement of $1.1\%$ from $42.7\%$ to $43.8\%$, which does not seem worth the additional annotation cost ($29.3$ versus $22.1$ seconds per image).

\smallsec{Random Annotators} Using the data from multiple annotators, we also ran experiments to estimate the effect of human variance on the accuracy of the model.  For each experiment, we randomly selected a different independent annotator to label each image. Three runs achieved $42.8$, $43.4$, and $43.8$ mIOU respectively, as compared to our original result of $42.7$ mIOU. This suggests that the variation in the location of the annotators' points does not significantly affect our results. This also further confirms that humans are predictable and consistent in pointing to objects~\cite{Clark05,Firestone14}. 

\smallsec{Random Points}
An interesting experiment is supervising with one point per class, but randomly sampled on the target object class using per-pixel supervised ground truth annotations (instead of asking humans to click on the object). This improved results over the human points by $3.4\%$, from $42.7\%$ to $46.1\%$. This is due to the fact that humans are predictable and consistent in pointing~\cite{Firestone14,Clark05}, which reduces the variety in point-level supervision across instances.

\begin{table*}[t]
\centering
\setlength\tabcolsep{1pt}
\resizebox{\textwidth}{!}{%
\begin{tabular}{p{2cm} |c cccccccccccccccccccc | c}
\textbf{Model}   & bg & aer & bic & bir & boa & bot & bus  & car  & cat  & cha & cow  & din & dog  & hor & mot & per & pot & she & sof & tra & tv & avg \\
\hline 


\IMG \newline  & 
60 & 
25 &	15	& 23 &	21	& 20	& 48	& 36	& 47	& 9	& 34 &	21 & 37 &	32	& 37	& 18	& 24	& 34 &	21	& 40	& 24 &
30 \\


\IMG+\OBJ  \newline& 
\textbf{79} 
& 42	& 20	& {\bf 39} &	33 &	17	& 34	& 39	& 45	& 10& 35	& 13	& 42	& 34	& 33 &	23	& 19 &	40	& 15 &	38	& 28 &
32 \\


\IMG+\pointclss  \newline& 
56 &
25 & 16 & 22  & 20 & 31 & 53 & 34 & \textbf{53} & 	8& \textbf{41} & \textbf{42} & 43 & \textbf{40} & 42 & 46 & 24 & 38 & \textbf{29} & 46 & 30 &
35 \\


\IMG+\pointclss+\OBJ& 
78 &
\textbf{49}	& \textbf{23}	& 37	& \textbf{37}	& \textbf{37}	& \textbf{57}	& \textbf{50} &	51	& \textbf{14} &	40	& 41 &	\textbf{50}	& 38 &	\textbf{51}	& \textbf{47}	& \textbf{31}	& \textbf{48} & 28 &	\textbf{49}	& \textbf{45} &
\textbf{43} \\

\hline 

\pointinsts \newline & 
79 &
49	& 21	& \textbf{40}	& 38 &	38	& 50	& 45	& 53	& 17 &	{\bf 43}	& 40	& 47	& {\bf 44}	& 51	& 51	& 22& 47	& {\bf 29} &	52	& 44 &
43 \\


\pointinsts \newline (weighted) & 
77 & 
48	& \textbf{23} 	& 38	& 36 &	38	& 57	& 52 &	52	& 13	& 42 & 	41 &	50 &	43	 & 52 & 46 & 31	& 49	& 28 &	50 &	44 &
43 \\


\pointclss \newline (3 annot.) & 
79 &
{\bf 50}	& 23	& 39 & 37	& 39	& 60	& 50	& 54	& 15	& 41 & 	\textbf{42}	& 49	& 42 &	52 & 50	& 29	& 49	& 29	& 49 & 44 &
44 \\


\pointclss \newline (random) & 
{\bf 80} & 
49	& 23 &	39	& {\bf 41}	& \textbf{46	} & {\bf 60} 	& {\bf 61}	& {\bf 56}	& {\bf 18} &	38	& 41	& {\bf 54} &	42	& {\bf 55}	& {\bf 57}	& \textbf{32}	& {\bf 51} &	26	& {\bf 55} &	\textbf{45} &
{\bf 46} \\

\hline
\end{tabular}}
\caption{Per-class segmentation accuracy (\%) on the PASCAL VOC 2012 validation set. (Top) Models trained with image-level, point supervision and (optionally) an objectness prior described in Section~\ref{sec:point-sup}. (Bottom) Models supervised with variations of point-level supervision described in Section~\ref{sec:point-variations}. 
}
\label{table:perclass}
\end{table*}

\subsection{Incorporating Stronger Supervision}
\label{sec:stronger}

\smallsec{Hybrid Approach with Points and Full Supervision} A fully supervised segmentation model achieves $58.3\%$ mIOU at a cost of $239.7$ seconds per image; recall that a point-level supervised model achieves $42.7\%$ at a cost of $22.4$ seconds per image. We explore the idea of combining the benefits of the high accuracy of full supervision with the low cost of point-level supervision. We train a hybrid segmentation model with a combination of a small number of fully-supervised images ($100$ images in this experiment), and a large number of point-supervised images (the remaining 10,482 images in PASCAL VOC 2012). This model achieves $53.1\%$ mIOU, a significant $10.4\%$ increase in accuracy over the \pointclss model, falling only $5.2\%$ behind full supervision. This suggests that the first few fully-supervised images are very important for learning the extent of objects, but afterwards, point-level supervision is quite effective at providing the location of object classes. Importantly, this hybrid model maintains a low annotation time, at  an average of only $24.5$ seconds per image: $(100 \times 239.7 + 10482 \times 22.4)/(100+10482) = 24.5$ seconds, which is $9.8$x cheaper than full supervision. We will further explore the tradeoffs between annotation cost and accuracy in Section~\ref{sec:budget}.

\smallsec{Squiggles}
Free-form squiggles are a natural extension of points towards stronger supervision. Squiggle-level supervision annotates a larger number of pixels: we collect an average of $502.7$ supervised pixels per image with squiggles, vs. $1.5$ with \pointcls. Like points, squiggles provide a nice tradeoff between accuracy and annotation cost. The squiggle-supervised model achieves $16.9\%$ higher mIOU than image-level labels and $6.4\%$ higher mIOU than \pointcls, at only $1.6-1.7$x the cost. However, squiggle-level supervision falls short of the hybrid approach on both annotation time and accuracy: squiggle-level takes a longer $35.2$ seconds compared to $24.5$ seconds for hybrid, and squiggle-level achieves only $49.1\%$ mIOU compared to the better $53.1\%$ mIOU with hybrid. This suggests that hybrid supervision combining large-scale point-level annotations with full annotation on a handful of images is a better annotation strategy than squiggle-level annotation.

\subsection{Segmentation Accuracy on a Budget}
\label{sec:budget}

\begin{figure}[t]
\begin{minipage}{0.48\linewidth}
\setlength{\tabcolsep}{3pt}
\centering
\begin{tabular}{|l|c|}
\hline
{\bf Supervision} & {\bf mIOU (\%)} \\
\hline
Full (883 imgs) & 22.1 \\
Image-level (10,582 imgs) & 29.8 \\
Squiggle-level (6,064 imgs) & 40.2 \\
Point-level (9,576 imgs) & {\bf 42.9} \\
\hline
\end{tabular}
\captionof{table}{Accuracy of models on the PASCAL VOC 2012 validation set given a fixed budget (and number of images annotated within that budget). Point-level supervision provides the best tradeoff between annotation time and accuracy. Details in Section~\ref{sec:budget}.}
\label{table:budget}
\end{minipage}\hfill
\begin{minipage}{0.48\linewidth}
\centering
\includegraphics[width=0.9\linewidth]{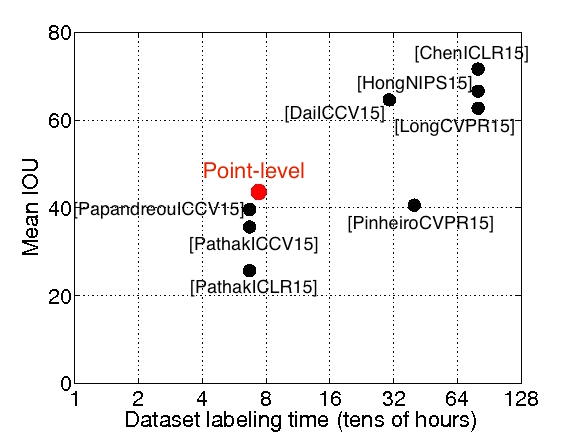}
\caption{Results without resource constraints on the PASCAL VOC 2012 \emph{test} set. The x-axis is log-scale.} 
\label{fig:testtime}
\end{minipage}
\end{figure}

\smallsec{Fixed Budget} Given a fixed annotation time budget, what is the right strategy to obtain the best semantic segmentation model possible? We investigate the problem by fixing the total annotation time to be the $10,582 \times (20.3) = 60$ hours that it would take to annotate all the $10,582$ training times with image-level labels. For each supervision method, we then compute the number of images $N$ that it is possible to label in that amount of time, randomly sample $N$ images from the training set, use them to train a segmentation model, and measure the resulting accuracy on the validation set. Table~\ref{table:budget} reports both the number of images $N$ and the resulting accuracy of fully supervised ($22.1\%$ mIOU), image-level supervised ($29.8\%$ mIOU), squiggle-level supervised ($40.2\%$ mIOU) and point-level supervised ($42.9\%$ mIOU) model. {\bf Point-level supervision outperforms the other types of supervision on a fixed budget}, providing an optimal tradeoff between annotation time and resulting segmentation accuracy. 

\smallsec{Comparisons to Others} For the rest of this section, we use a model trained on all 12,031 training+validation images and evaluate on the PASCAL VOC 2012 \emph{test} set (as opposed to the validation set above) to allow for fair comparison to prior work. Point-level supervision (\IMG + \pointcls + \OBJns) obtains $43.6\%$ mIOU on the test set. Fig.~\ref{fig:testtime} shows the tradeoffs between annotation time and accuracy of different methods, discussed below.

\smallsec{Unlimited Budget (Strongly Supervised)} We compare both the annotation time and accuracy of our point-supervised \pointclss model with published techniques with much larger annotation budgets, as a reference for what might be achieved by our method if given more resources. Long \emph{et al.}~\cite{Long15} reports $62.2\%$ mIOU, 
Hong \emph{et al.}~\cite{hong2015decoupled} reports $66.6\%$ mIOU, and 
Chen \emph{et al.}~\cite{DeepLab-strong} reports $71.6\%$ mIOU, 
but in the fully supervised setting that requires about 800 hours of annotation, an order of magnitude more time-consuming than point supervision. Future exploration will reveal whether point-level supervision would outperform a fully supervised algorithm given 800 annotation hours of data.

\smallsec{Small Budget (Weakly Supervised)} We also compare to weakly supervised published results. Pathak ICLR \emph{et al.}~\cite{Pathak15} achieves $25.7\%$ mIOU, Pathak ICCV \emph{et al.}~\cite{pathak2015constrained} achieves $35.6\%$ mIOU, and Papandreou \emph{et al.}~\cite{papandreouweakly} achieves $39.6\%$ mIOU with only image-level labels requiring approximately $67$ hours of annotation on the 12,301 images (Section~\ref{sec:human}).  
Pinheiro et al.~\cite{Pinheiro15} achieves $40.6\%$ mIOU but with $400$ hours of annotations.\footnote{\cite{Pinheiro15} trains with only image-level annotations but adds 700,000 additional positive ImageNet images and 60,000 background images. We choose not to count the 700,000 freely available images but the additional 60,000 background images they annotated would take an additional $60,000 \times 20 $ classes $\times 1$ second = 333 hours. The total annotation time is thus $333 + 67 = 400$ hours.}
We improve in accuracy upon all of these methods and achieve $43.6\%$ with point-level supervision requiring about 79 annotation hours.
Note that our baseline model is a significantly simplified version of~\cite{Pathak15,papandreouweakly}. Incorporating additional features of their methods is likely to further increase our accuracy at no additional cost.

\smallsec{Size constraint} Finally, we compare against the recent work of~\cite{pathak2015constrained} which trains with image-level labels but incorporates an additional bit of supervision in the form of object size constraints. They achieve $43.3\%$ mIOU (omitting the CRF post-processing), on par with $43.6\%$ using point-level supervision. This size constraint should be fast to obtain although annotation times are not reported. These two simple bits of supervision (point-level and size) are complementary and may be used together effectively in the future.

\section{Conclusions}
We propose a new time-efficient supervision approach for semantic image segmentation based on humans pointing to objects. We show that this method enables training more accurate segmentation models than other popular forms of supervision when given the same annotation time budget. In addition, we introduce an objectness prior directly in the loss function of our CNN to help infer the extent of the object. We demonstrated the effectiveness of our approach by evaluating on the PASCAL VOC 2012 dataset. We hope that future large-scale semantic segmentation efforts will consider using the point-level supervision we have proposed, building upon our released dataset and annotation interfaces.

\subsubsection*{Acknowledgments}
We would like to thank Evan Shelhamer for helping us set up the baseline model of~\cite{Long15}, as well as all the other Caffe developers. We also thank Lamberto Ballan, Michelle Greene, Anca Dragan, and Jon Krause. 


V. Ferrari was supported by the ERC Starting Grant VisCul. L. Fei-Fei was supported by an ONR-MURI grant. GPUs were graciously donated by NVIDIA. 

\bibliographystyle{splncs}  
\bibliography{egbib}
\end{document}